\documentclass{article}

     \usepackage[final,nonatbib]{neurips_2020}

\usepackage[utf8]{inputenc} 
\usepackage[T1]{fontenc}    
\usepackage{hyperref}       
\usepackage{url}            
\usepackage{booktabs}       
\usepackage{amsfonts}       
\usepackage{amsmath}
\usepackage{amsthm}
\usepackage{nicefrac}       
\usepackage{microtype}      
\usepackage{graphicx}
\usepackage{epstopdf}
\usepackage{wrapfig}
\usepackage{subcaption}
\usepackage{algorithm}
\usepackage{algorithmic}

\newcommand{\R}{\mathbb{R}}

\newcommand{\N}{\mathbb{N}}

\newcommand{\norm}[1]{\lVert{#1}\rVert}

\newcommand{\bmat}[1]{\begin{bmatrix}#1\end{bmatrix}}
\newcommand{\secspace}{\vspace{-4mm}}
\newcommand{\subsecspace}{\vspace{-3mm}}
\newcommand{\proofspace}{\vspace{-4mm}}
\newcommand{\presecspace}{\vspace{-3mm}}

\newtheorem{thm}{Theorem}
\newtheorem{prop}[thm]{Proposition}
\newtheorem{lem}[thm]{Lemma}
\newtheorem{defn}[thm]{Definition}
\newtheorem{corr}[thm]{Corollary}
\newcommand{\ip}[2]{\langle #1, #2 \rangle}
\newcommand{\mcl}[1]{\mathcal{#1}}

\title{A New Algorithm for Tessellated Kernel Learning}

\author{
  Brendon K.~Colbert \\
  Arizona State University\\
  Tempe, AZ 85287 \\
  \texttt{brendon.colbert@asu.edu} \\
  \And
  Matthew M. Peet \\
  Arizona State University \\
  Tempe, AZ 85287 \\
  \texttt{mpeet@asu.edu} \\
}

\begin{document}

\maketitle

\begin{abstract}
The accuracy and complexity of machine learning algorithms based on kernel optimization are limited by the set of kernels over which they are able to optimize. An ideal set of kernels should: admit a linear parameterization (for tractability); be dense in the set of all kernels (for robustness); be universal (for accuracy). The recently proposed Tesselated Kernels (TKs) is currently the only known class which meets all three criteria. However, previous algorithms for optimizing TKs were limited to classification and relied on Semidefinite Programming (SDP) - limiting them to relatively small datasets. By contrast, the 2-step algorithm proposed here scales to 10,000 data points and extends to the regression problem. Furthermore, when applied to benchmark data, the algorithm demonstrates significant improvement in performance over Neural Nets and SimpleMKL with similar computation time.
\end{abstract}

\section{Introduction} \secspace
Kernel methods for classification and regression (and Support Vector Machines (SVMs) in particular) require selection of a kernel. Kernel Learning (KL) algorithms such as those found in~\cite{xu2010simple,sonnenburg2010shogun,yang2011efficient} automate this task by finding the kernel, $k \in \mcl K$ which optimizes an achievable metric such as the soft margin (for classification). The set of kernels, $k \in \mcl K$, over which the algorithm can optimize, however, strongly influences the performance and robustness of the resulting classifier or predictor.

To understand how the choice of $\mcl K$ influences performance and robustness, three properties were proposed in~\cite{JMLR} to characterize the set $\mcl K$ - tractability, density, and universality. Specifically, $\mcl K$ is tractable if $\mcl K$ is convex (or, preferably, a linear variety) - implying the KL problem is solvable using, e.g.~\cite{rakotomamonjy_2008,jain_2012,lanckriet_2004,qiu2005multiple,gonen2011multiple}. The set $\mcl K$ has the density property if, for any $\epsilon>0$ and any positive kernel, $k^*$ there exists a $k \in \mcl K$ where $\norm{k-k^*}\le \epsilon$. The density property implies the kernel will perform well on untrained data (robustness or generalizability). The set $\mcl K$ has the universal property if any $k \in \mcl K$ is universal - ensuring the classifier/predictor will perform arbitrarily well on large sets of training data.

In \cite{JMLR}, the Tessellated Kernels (TKs) were shown to have all 3 properties, the first known such class of kernels. This work was based on a general framework for using positive matrices to parameterize positive kernels (as opposed to positive kernel matrices as in~\cite{lanckriet_2004,qiu2005multiple,ni2006learning}). Unfortunately, however, the algorithms proposed in~\cite{JMLR} were either based on SemiDefinite Programming (SDP) (thereby limiting the amount of training data) or used a randomized linear basis for the kernels (implying loss of density).
Thus, while the algorithms in~\cite{JMLR} outperformed all other methods (including deep learning) as measured by Test Set Accuracy (TSA), the computation times were not competitive. Furthermore, the results in~\cite{JMLR} did not encompass the problem of regression.

In this paper, we extend the TK framework proposed in~\cite{JMLR} to the problem of regression. The KL problem in regression has been studied using SDP in~\cite{qiu2005multiple,ni2006learning} and Quadratic Programming (QP) in e.g.~\cite{rakotomamonjy_2008,jain_2012}. However, neither of these previous works considered a set of kernels with both the tractability and the density property. By generalizing the Tessellated KL framework proposed in~\cite{JMLR} to the regression problem, we demonstrate significant increases in performance, as measured by Mean Square Error (MSE), and when compared to the results in~\cite{rakotomamonjy_2008,jain_2012,qiu2005multiple}.

In addition, we show that the SDP-based algorithm~\cite{JMLR} for classification, and extended here to regression, can be decomposed into primal and dual sub-problems, $OPT\_A$ and $OPT\_P$ - similar to the approach taken in~\cite{rakotomamonjy_2008,jain_2012}. Furthermore, we show that $OPT\_P$ (an SDP) admits an analytic solution using the Singular Value Decomposition (SVD) - an approach which allows us to consider higher dimensional feature spaces and more complex TKs. In addition, $OPT\_A$ is a convex QP and may be solved efficiently with achieved complexity which scales as $O(m^{2.16})$ where $m$ is the number of data points. We use a two-step algorithm on $OPT\_A$ and $OPT\_P$ and show that termination at $OPT\_A=OPT\_P$ is equivalent to global optimality. The resulting algorithm, then, does not require the use of SDP and, when applied to several standard test cases, is shown to retain the favorable TSA of~\cite{JMLR} for classification, while offering improved MSE for regression, and competitive computation times as compared to other KL and deep learning algorithms. \presecspace
\section{An Ideal Set of Kernels for KL in Classification and Regression} \label{sec:Ideal} \secspace
Consider a generalized representation of the KL problem, which encompasses both classification and regression where (using the representor theorem~\cite{scholkopf2001generalized}) the learned function is of the form $f_{\alpha,k}(z) = \sum_{i=1}^m \alpha_i k(x_i,z)$.\vspace{-2mm}
\begin{equation} \label{eqn:OPT}
\min_{k  \in \mcl{K}} \min_{\alpha \in \R^m,b} \norm{f_{\alpha,k}}^2 + C \sum\nolimits_{i=1}^m l(f_{\alpha,k},b)_{y_i,x_i}
\end{equation}
Here $\norm{f_{\alpha,k}}_X = \sum_{i=1}^m \sum_{j=1}^m \alpha_i \alpha_j k(x_i,x_j)$ and $l(f_{\alpha,k},b)_{y_i,x_i}$ is the loss function and is defined for SVM binary classification and SVM regression as $l^c(f_{\alpha,k},b)_{y_i,x_i}$ and $l^r(f_{\alpha,k},b)_{y_i,x_i}$, respectively, where
\[
l^c(f_{\alpha,k},b)_{y_i,x_i} \hspace{-1mm}=  \max \{  0, 1 - y_i (f_{\alpha,k}(x_i) - b) \}~~\text{and}~~ l(f_{\alpha,k},b)^r_{y_i,x_i} \hspace{-1mm}=  \max \{  0, |y_i - (f_{\alpha,k}(x_i) - b)|- \epsilon \}.
\]



The properties of the classifier/predictor, $f_{\alpha,k}$, resulting from Optimization Problem~\ref{eqn:OPT} will depend on the properties of the set $\mcl K$, which is presumed to be a subset of the convex cone of all positive kernels. To understand how $\mcl K$ influences the tractability of the optimization problem and the resulting fit, we consider three properties of the set, $\mcl K$.
%

\subsection{Tractability} \subsecspace

We say a set of kernel functions, $\mcl K$, is tractable if it can be represented using a countable basis.
\begin{defn} \label{Tractability}
The set of kernels $\mcl K$ is \textit{\textbf{tractable}} if there exist a countable set $\{G_{i}(x,y)\}_{i}$ such that, for any $k \in \mcl K$, there exists $N_G \in \N$ where $k(x,y)=\sum_{i=1}^{N_G} v_i G_i(x,y)$ for some $v \in \R^{N_G}$.
\end{defn}
Note the $G_i(x,y)$ need not be positive kernel functions. The tractable property is required for the KL problem to be tractable using algorithms for convex optimization.

\subsection{Universality}\label{Universal_Motivation} \subsecspace
Universal kernel functions always have positive definite (full rank) kernel matrices, implying that for arbitrary data $\{y_i,x_i\}_{i=1}^m$, there exists a function $f(z) = \sum_{i=1}^m \alpha_i k(x_i,z)$, such that $f(x_j) = y_j$ for all $j= 1,..,m$. Conversely, if a kernel is not universal, then exists a data set $\{x_i,y_i\}_{i=1}^m$ such that for any $\alpha \in \R^m$, there exists some $j\in \{1,\cdots,m\}$ such that $f(y_j) \neq \sum_{i=1}^m \alpha_i k(x_i,x_j)$. This ensures that SVMs using universal kernels can always benefit from additional training data, whereas non-universal kernels may saturate.

\begin{defn}
A kernel $k:X \times X \rightarrow \R$ is said to be \textit{universal} on the compact metric space $X$ if it is continuous and there exists an inner-product space $\mcl W$ and feature map, $\Phi : X \rightarrow \mcl W$ such that $k(x,y)=\ip{\Phi(x)}{\Phi(y)}_{\mcl W}$ and where the unique Reproducing Kernel Hilbert Space (RKHS), $\mcl H:=\{f\;:\;f(x)=\ip{v}{\Phi(x)},\; v \in \mcl W\}$
with associated norm $\norm{f}_{\mcl H} :=\inf_{v}\{\norm{v}_{\mcl W}\;:\; f(x)=\ip{v}{\Phi(x)}\}$ is dense in $\mcl C(X):=\{f \,: \, X \rightarrow \R \;:\; $f$\, \text{ is continuous}\}$ where $\norm{f}_{\mcl C}:=\sup_{x\in X} |f(x)|$.
\end{defn}
The following definition extends the universal property to a set of kernels.
\begin{defn}
A set of kernel functions $\mcl K$ has the universal property if every kernel function $k \in \mcl K$ is universal.
\end{defn}

\subsection{Density} \subsecspace
The third property is density which distinguishes the TK class from other sets of kernel functions with the universal property. For instance consider a set containing a single Gaussian kernel function - which is clearly not ideal for kernel learning.  The set containing a single Gaussian is tractable (it has only one element) and every member of the set is universal. However, it is not dense.

Considering SVM for classification, the KL problem determines the kernel $k \in \mcl K$ for which we may obtain the maximum separation in the kernel-associated feature space. Increasing this separation distance makes the resulting classifier more robust (generalizable)~\cite{boehmke2019hands}. The density property, then, ensures that the resulting KL algorithm will be maximally robust (generalizable) in the sense of separation distance.

Likewise, considering SVMs for regression, the KL problem finds the kernel $k\in \mcl K$ which permits the ``flattest''~\cite{smola2004tutorial} function in feature space. In this case, the density property ensures that the resulting KL algorithm will be maximally robust (generalizable) in the sense of flatness.

These arguments motivate the following definition of the pointwise density property.
\begin{defn}
The set of kernels $\mcl K$ is said to be \textbf{pointwise dense} if for any positive kernel, $k^*$, any set of data $\{x_i\}_{i=1}^m$, and any $\epsilon>0$, there exists $k \in \mcl K$ such that $\norm{k(x_i,x_j)-k^*(x_i,x_j)}\le \epsilon$.
\end{defn}
\presecspace
\section{A General Framework for Representation of Tractable Kernel Sets} \label{sec:4} \secspace
Here we define a framework for constructing classes of tractable positive kernel functions and illustrate this approach on the class of General Polynomial Kernels. 

\begin{prop}\label{prop:kernel}
Let $N$ be any bounded measurable function $N: X \times Y \rightarrow \R^q$ and $P \in \R^{q \times q}$ be a positive semidefinite matrix $P\geq 0$. Then
\begin{equation}
k(x,y)=\int_{X} N(z,x)^T P N(z,y) dz\label{eqn:kernel}\vspace{-2mm}
\end{equation}
is a positive kernel function.
\end{prop}
The proof for Proposition~\eqref{prop:kernel} may be found in \cite{JMLR}.

\begin{lem}\label{lem:tractable}
Let $N$ be any bounded measurable function $N: X \times Y \rightarrow \R^q$ on compact $X$ and $Y$. Then the set of kernel functions
\begin{equation}
{\mcl K} := \left\{ k ~|~ k(x,y)=\int_{X} N(z,x)^T P N(z,y) dz, ~ P \geq 0 \right\} \qquad \text{is tractable.}\label{eqn:tractable}
\end{equation}
\end{lem}
For a given $N$, the map $P \mapsto k$ is linear. Specifically,
\[
k(x,y)=\sum\nolimits_{i=1}^q \sum\nolimits_{j=1}^q P_{i,j} G_{i,j}(x,y)~~\text{where}~~ G_{i,j}(x,y)=\int_{X}N_{i}(z,x)N_{j}(z,y)dz.
\]
and thus by Definition~\ref{Tractability} $\mcl K$ is tractable.

In Subsection~\ref{subsec:GPK} we apply this framework to obtain Generalized Polynomial Kernels. In Subsection~\ref{subsec:TK}, we use the framework to obtain the TK class.

\subsection{The Class of General Polynomial Kernels is Tractable}\label{subsec:GPK} \subsecspace
The class of General Polynomial Kernels (GPKs) is defined as the set of all polynomials, each of which is a positive kernel.
\begin{align}
 \mcl K_P &:= \{ k \in \R[x,y]\;:\; k \text{ is a positive kernel}  \} \label{polyKernel}
\end{align}
The GPK class is not universal, but is tractable, as per the following lemma.
\begin{lem}
$\mcl K_P$ is tractable.
\end{lem} \proofspace

\begin{proof}
Let $Z_d: \R^n \rightarrow \R^q$ be the vector of monomials of degree $d$ or less.  From~\cite{JMLR}, we have that a polynomial $k$ of degree $2d$ is a positive polynomial kernel if and only if there exists some $P\geq 0$ such that $k(x,y)=Z_d(x)^T P Z_d(y)$. Now for any finite-dimensional subset of $\mcl K_P$, let $d$ be the maximum degree over this subset and define $N(z,y)=Z_d(y)$. Then Lemma~\ref{lem:tractable} implies that $\mcl K_P$ is tractable.
\end{proof}

This lemma implies that a representation of the form of Equation~\eqref{eqn:kernel} is necessary and sufficient for a GPK to be positive. For convenience, we denote the set of GPK kernels of degree $d$ or less as follows~\cite{recht2006convex}.
\begin{align}
 \mcl K_P^d &:= \{ k\;:\; k(x,y) = Z_d(x)^T P Z_d(y) \; : \; P \geq 0  \} \label{polyKernel} 
\end{align}
\presecspace

\section{Tessellated Kernels: Tractable, Dense and Universal} \label{sec:TK} \secspace
In this section, we define the class of TK kernels and show it is tractable, dense, and universal.
\subsection{Tessellated Kernels}\label{subsec:TK} \subsecspace
Again, let $Z_d: \R^n  \times \R^n \rightarrow \R^q$ be the vector of monomials of degree $d$. Define $\mathbf I$, the indicator function for the positive orthant, and the following choice of $N: \R^n \times \R^n \rightarrow \R^{2q}$ as \vspace{-2mm}
\begin{equation}
\mathbf I(z) = \begin{cases}
    1       & \quad z \ge 0\\
    0  & \quad \text{otherwise,}\\
\end{cases} \quad ~~ \text{and} ~~ \quad N^d_{T}(z,x) = \bmat{Z_d(z,x)\mathbf I(z-x) \\ Z_d(z,x)\mathbf I(x-z) } \label{eqn:N}
\end{equation}
\vspace{-2mm}
where $z\ge 0$ means $z_i \ge 0$ for all $i$.
%
We now define the set of TK kernels for $a<b \in \R^n$ as
\[
\mcl K^d_T:=\left\{k\;:\; k(x,y) = \int_{a}^b N_T^d(z,x)^T \; P\;  N^d_{T}(z,y) dz,\; P\ge 0\right\},~ \mcl K_T:=\{k\;:\; k\in \mcl K_T^d,\; d \in \N\}.
\]
Kernels in the TK class are ``Tessellated'' in the sense that each datapoint defines a vertex which bisects each dimension of the domain of the resulting classifier/predictor - resulting in a tessellated partition of the feature space.


\subsection{The Set of TK Kernels is Tractable} \subsecspace
The class of TK kernels is prima facie in the form of Eqn.~\eqref{eqn:tractable} in Lemma~\ref{lem:tractable} and hence is tractable.

However, we will expand on this result by specifying the basis for the set of TK kernels, which will then be used in Section~\ref{sec:2step}.


\begin{corr}\label{corr}
Suppose that for $a<b \in \R^n$, and $d \in\N$. We define the finite set $D_d:=\{(\delta,\lambda)\in \N^{2n} : \norm{(\delta,\lambda)}_1 \le d\}$.  Let $\{ [\delta_i, \gamma_i] \}_{i=1}^q \subseteq D_d$ be some ordering of $D_d$ and define $Z_d(x,z)_j = x^{\delta_j} z^{\gamma_j}$ where $z^{\delta_j} x^{\gamma_j}:=\prod_{i=1}^n z_i^{\delta_j,i}x_i^{\gamma_j,i}$. Now let $k$ be as defined in Eqn.~\eqref{eqn:kernel} for some $P>0$ and where $N$ is as defined in Eqn.~\eqref{eqn:N}. If we partition $P = \bmat{Q & R\\ R^T & S}$ then we have,\vspace{-2mm}
\[
k(x,y)=\sum\nolimits_{i,j=1}^q Q_{i,j} g_{i,j}(x,y) + R_{i,j}t_{i,j}(x,y) + R^T_{i,j}t_{i,j}(y,x)  + S_{i,j}  h_{i,j}(x,y)
\]
where $g_{i,j},t_{i,j},h_{i,j}:\R^{2n}\rightarrow \R$ are defined as
\begin{align}\label{g}
g_{i,j}(x,y) &:= x^{\delta_i}y^{\delta_j} T(p^*(x,y),b,\gamma_{i,j} + \mathbf{1} ), \; t_{i,j}(x,y) := x^{\delta_{i}}y^{\delta_{j}} T(x,b,\gamma_{i,j} + \mathbf{1}  ) - g_{i,j}(x,y), \; \text{and}\notag \\
 h_{i,j}(x,y) &:= x^{\delta_{i}}y^{\delta_{j}} T(a,b,\gamma_i + \gamma_j + \mathbf{1}  ) - g_{i,j}(x,y) - t_{i,j}(x,y) - t_{i,j}(y,x), \notag
\end{align}
where $\mathbf{1} \in \N^n$ is the vector of ones, $p^*:\R^{2n}\rightarrow \R^n$ is defined elementwise as
$p^*(x,y)_i = \max \{x_i,y_i \}$, and $T:\R^n \times \R^n \times \N^n\rightarrow \R$ is defined as \[T(x,y,\zeta) = \prod\nolimits_{j=1}^n \left(\hspace{2mm} \frac{y_j^{\zeta_j}}{\zeta_j}-\frac{x_j^{\zeta_j}}{\zeta_j}\right).
\]
\vspace{-5mm}
\end{corr}
The proof of Corollary~\ref{corr} can be found in~\cite{JMLR}.
\subsection{The TK Class is Dense} \subsecspace
The density property differentiates the set of TK kernels from other sets of kernel functions (e.g. a linear combination of Gaussian kernels of fixed bandwidths).

From \cite{JMLR} we have that the set of TK kernels satisfies the pointwise density property.
\begin{thm} \label{thm:TessellatedOptimal}
For any kernel matrix $K^*$ and any finite set $\{x_i\}_{i=1}^m$, there exists a $d \in\N$ and $k \in\mcl K_T^d$ such that if $K_{i,j} = k(x_i,x_j)$, then $K=K^*$.
\end{thm}

\begin{wrapfigure}{R}{0.32\textwidth}
\vspace{-1mm}
 \begin{center}
\includegraphics[trim= 20 0 50 20, clip, width=.32\textwidth]{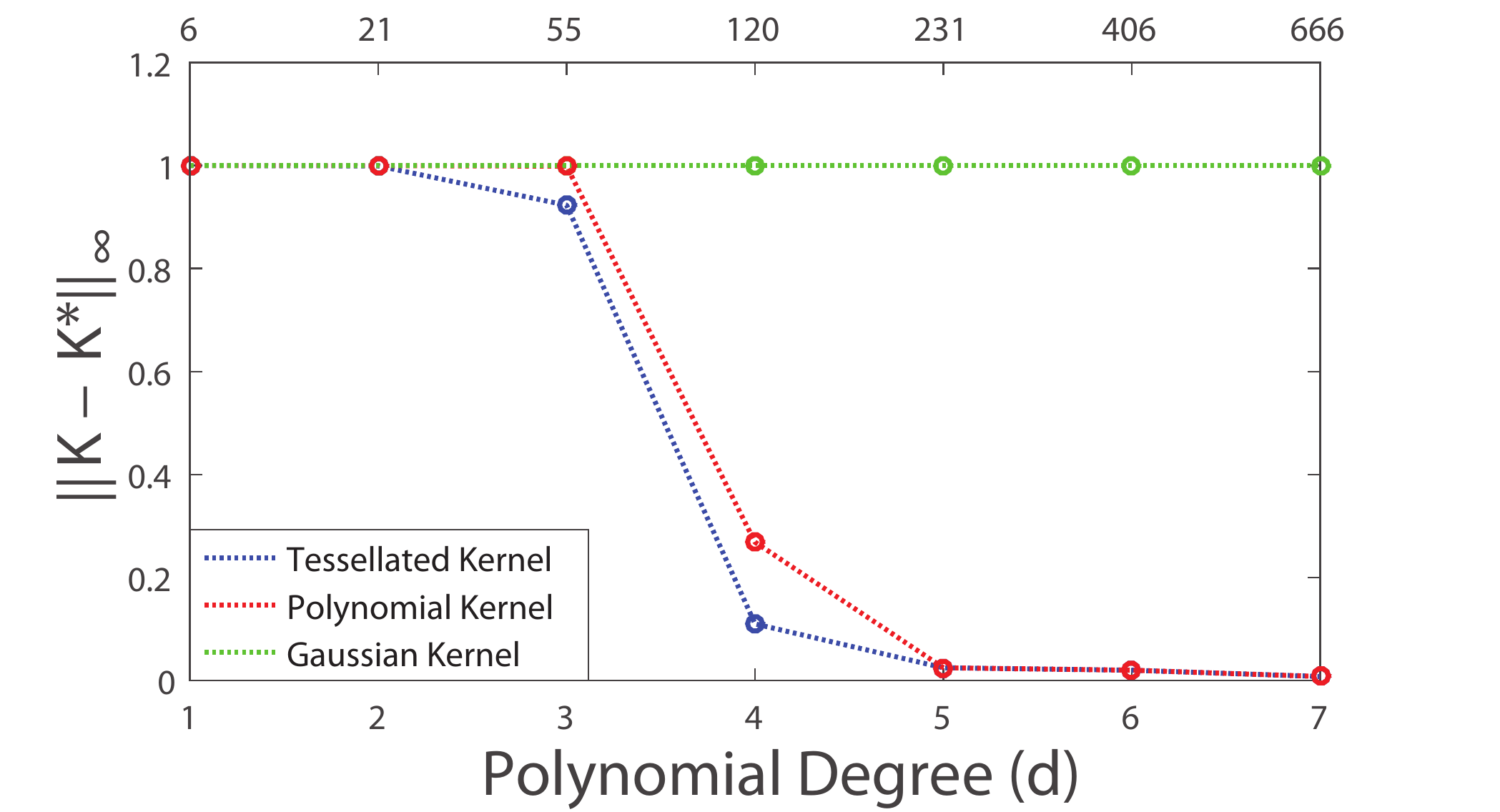}
  \end{center}\vspace{-3mm}
        \caption{The achieved objective ($\norm{K-K^*}_\infty$) of Optimization Problem~\ref{opt:infnorm} for TK and GPK of degree $d$; and $m$ Gaussians with bandwidths in $[.01,10]$.  The number of bandwidths is selected so that the number of decision variables match in the Gaussian and TK cases.
        }\label{fig:density}
\vspace{-7mm}
  \end{wrapfigure}
  
In~\cite{JMLR} an analytical solution, $K^*$, was found for the optimal trace-constrained kernel matrix that maximized the separation distance between two classes of points in the feature space. It was shown in this work that when $\{y_i\}$ has an equal number of positive and negative labels, $K^*$ contains an equal number of positive and negative elements - illustrating the importance of using kernels which are not pointwise positive (Gaussians are pointwise positive).

To illustrate the density property, then, we show how optimal GPK and TK kernels yield kernel matrices which approximate the analytic solution, $K^*$, of the optimal kernel matrix problem for a given set of data $\{x_i\}$ and labels $\{ y_i \}$, while Gaussian kernels do not. Specifically, we consider the following optimization problem. \vspace{-2mm}
\begin{equation}
\min_{k \in \mcl K} \norm{K-K^*}_\infty~~~ s.t. ~~K_{i,j} = k(x_i,x_j) \label{opt:infnorm}
\end{equation}
In these problems, the sets $\mcl K$ will be: $\mcl K_G^\gamma$ - the sum of $N$ Gaussians with bandwidths $\gamma_i$; $\mcl K_P^d$ - the GPKs of degree $d$; and $\mcl K_T^d$  - the TK kernels of degree $d$. More precisely, for bandwidths $\gamma\in \R^{N}$, we define $\mcl K^\gamma_G := \left\{k\;:\; k(x,y) = \sum_{i = 1}^N \mu_i \text{e}^\frac{||x-y||_2^2}{\gamma_i} \; : \; \mu_i > 0  \right\}.$


Consider a spiral data set with 20 samples, using equal numbers of positive and negative labels.
Fig.~\ref{fig:density} shows the achieved objective value of Problem~\eqref{opt:infnorm} for $\mcl K^\gamma_G$, $\mcl K_P^d$, and $\mcl K^d_T$ as a function of the number of bandwidths (top $x$ axis - $N$ in $\mcl K^\gamma_G$), polynomial degree (bottom $x$ axis - $d$ in $\mcl K_P^d$, and $\mcl K^d_T$). The $x$-axes of the plots are scaled to show equal numbers of decision variables. As expected,
the case $\mcl K=\mcl K_\gamma^G$ saturates with an objective value significantly larger than the lower bound.  The cases $\mcl K=\mcl K_P^d$ and $\mcl K=\mcl K^d_T$, meanwhile have almost no error at degree $d=7$.\presecspace

\subsection{TK Kernels are Universal} \subsecspace
Finally we discuss the universality property of the class of TK kernels which ensures that every TK function can fit the training data well.

The following theorem from \cite{JMLR} shows that any TK kernel with $P > 0$ is necessarily universal.
\begin{thm} \label{thm:universal}
Suppose $k$ is as defined in Eqn.~\eqref{eqn:kernel} for some $P>0$, $d \in\N$ and $N$ as defined in Eqn.~\eqref{eqn:N}. Then $k$ is universal for $a<b \in \R^n$.
\end{thm}
This theorem implies that even if we use the subset of TK kernels defined by $d=0$, this subset is still universal.  \presecspace

\section{A New Algorithm for KL in Classification and Regression using TKs}\label{sec:2step} \secspace
In this section, we express the KL optimization problem for both classification and regression and break this optimization problem into two sub-problems which allow us to express the problem in primal and dual form. For convenience, we define the feasible sets for the sub-problems as
\[
\mcl X:=\{P \in \R^{q \times q}\;:\; \text{trace}(P) = q,\; P > 0\}\vspace{-2mm}
\]
\[
\mcl Y_c:=\{\alpha \in \R^m\; : \; \sum\limits_{i=1}^m \alpha_iy_i = 0,\; 0 \leq \alpha_i \leq C \},\quad
\mcl Y_r:=\{\alpha \in \R^m\;:\; \sum\limits_{i=1}^m \alpha_i = 0, \; \alpha_i \in [-C, C]\}.
\]
The common part of the objective is
\begin{equation} \label{OalphaP}
O(\alpha,P):= -\frac{1}{2} \sum\nolimits_{i=1}^m\sum\nolimits_{j=1}^m \alpha_i \alpha_j \int_{a}^b N_T^d(z,x_i)^T P N^d_{T}(z,y_j) dz,
\end{equation}
while the unique parts of the objective are
\[
\kappa_c(\alpha):=\sum\nolimits_{i=1}^m \alpha_i\qquad\text{and}\qquad 
\kappa_r(\alpha):=-\epsilon \sum\nolimits_{i=1}^m |\alpha_i| + \sum\nolimits_{i=1}^m y_i\alpha_i.
\]
Then the KL optimization problem ($OPT$) for TK kernels ($\odot$ being elementwise multiplication) is as follows for classification and regression, respectively.
\begin{align*}
OPT:=\min_{P \in \mcl X} \max_{~\alpha \in \mcl Y_c}  \quad O(\alpha \odot y,P) + \kappa_c(\alpha), \quad \text{and} \quad OPT:=\min_{P \in \mcl X} \max_{~\alpha \in \mcl Y_r}  \quad O(\alpha,P) + \kappa_r(\alpha).
\end{align*}

\textbf{Primal Formulation:} We can formulate the primal problem ($OPT_P$) as
\begin{align}  \label{KernelSVC}
OPT_P=\min_{P \in \mcl X} \max_{\alpha \in \mcl Y} O(\alpha, P)+\kappa_r(\alpha)\; \text{or}\;O(\alpha\odot y, P)+\kappa_c(\alpha) =\min_{P \in \mcl X} ~~ OPT\_A(P) 
\end{align}
where for classification and regression, respectively,
\[
OPT\_A(P):=\max_{\alpha \in \mcl Y_c} \;\; O(\alpha\odot y, P)+\kappa_c(\alpha), \quad \text{and} \quad OPT\_A(P):=\max_{\alpha \in \mcl Y_r}\;\; O(\alpha, P)+\kappa_r(\alpha).
\]

\textbf{Dual Formulation:} Alternatively, we have the dual formulation ($OPT_D$).
\begin{align}  \label{KernelSVC}
OPT_D=\max_{\alpha \in \mcl Y} OPT\_P( \alpha) 
\end{align}
where $\mcl Y=\mcl Y_c$ for classification and $\mcl Y=\mcl Y_r$ regression. Likewise, for classification and regression, respectively,
\[
OPT\_{P}(\alpha):=\min_{P\in \mcl X}  O(\alpha \odot y, P)+\kappa_c(\alpha) \quad \text{and} \quad OPT\_{P}(\alpha):=\min_{P\in \mcl X}  O(\alpha, P) +\kappa_r(\alpha).
\]


\begin{lem}
For $\alpha \in \mcl Y$, $P \in \mcl X$, $OPT\_A(P)=OPT\_P(\alpha)$  if and only if: $\{\alpha,P\}$ solve $OPT$; $P$ solves $OPT_P$; and $\alpha$ solves $OPT_D$.
\end{lem} \proofspace
\begin{proof}
For any minmax optimization problem with objective function $\phi$, we have
\[d^* = \max_{\alpha \in \mcl Y} \min_{P \in \mcl X} \phi(P,\alpha) \leq   \min_{P \in \mcl X}\max_{\alpha \in \mcl Y}  \phi(P,\alpha) = p^*, \]
and strong duality holds ($p^*-d^* = 0$) if $\mcl X$ and $\mcl Y$ are both convex and one is compact, $\phi(\cdot,\alpha)$ is convex for every $\alpha \in \mcl Y$ and $\phi(P,\cdot)$ is concave for every $P \in \mcl X$, and the function $\phi$ is continuous \cite{fan1953minimax}. In our case, these conditions hold for both classification and regression where $\phi(P,\alpha)=O(\alpha, P)+\kappa_r(\alpha)\; \text{or}\;O(\alpha\odot y, P)+\kappa_c(\alpha)$. Hence if $\alpha^*$ solves $OPT\_P$ and $P^*$ solves $OPT\_A$, then $\{\alpha^*, P^*\}$ solves $OPT$ and\vspace{-1.5mm}
\[
OPT\_P( \alpha^*)=\max_{\alpha \in \mcl Y} OPT\_P( \alpha)=\min_{P \in \mcl X} ~~ OPT\_A(P)=OPT\_A(P^*).\vspace{-2.5mm}
\]
Conversely, suppose $\alpha \in \mcl Y$, $P \in \mcl X$, then \vspace{1mm}
\begin{align*}\small
OPT\_P(\alpha) \le \max_{\alpha \in \mcl Y}\; OPT\_P(\alpha) &= OPT\_P( \alpha^*)\\[-2mm]
& = OPT\_A(P^*)= \min_{P \in \mcl X} ~~ OPT\_A(P) \le OPT\_A(P).\\[-6mm]
\end{align*}
Hence if $OPT\_A(P)=OPT\_P(\alpha)$, then $OPT\_A(P)=OPT\_A(P^*)=OPT\_P(\alpha^*)=OPT\_P(\alpha)$ and hence $P$ and $\alpha$ solve $OPT\_A$ and $OPT\_P$, respectively.
\end{proof}\presecspace
We propose Algorithm~\ref{TKL} as a two-step iterative algorithm for solving Optimization Problem~\eqref{KernelSVC}.
\begin{wrapfigure}{R}{0.5\textwidth}
\vspace{-15mm}
\begin{minipage}{0.5\textwidth}
\begin{algorithm}[H]
\begin{algorithmic}
\STATE \texttt{Initialize} $P=I$;
\WHILE{$OPT\_P(\alpha_k)-OPT\_A(P_k)>\epsilon$}
\STATE $\alpha_{k+1}= OPT\_A(P_k)$
\STATE $P_{k+1}= \frac{P_K + t OPT\_P(\alpha_{k+1})}{1+t}$ (select $t$ using line search)
\STATE $k= k+1$
\ENDWHILE
\end{algorithmic}
\caption{Two Step TKL} \label{TKL}
\end{algorithm}
 \end{minipage}
\vspace{-4mm}
  \end{wrapfigure}

\begin{figure}[t]
    \centering
    \begin{subfigure}[t]{0.23\textwidth}
        \centering
\includegraphics[trim= 20 0 50 20, clip, width=.95\textwidth]{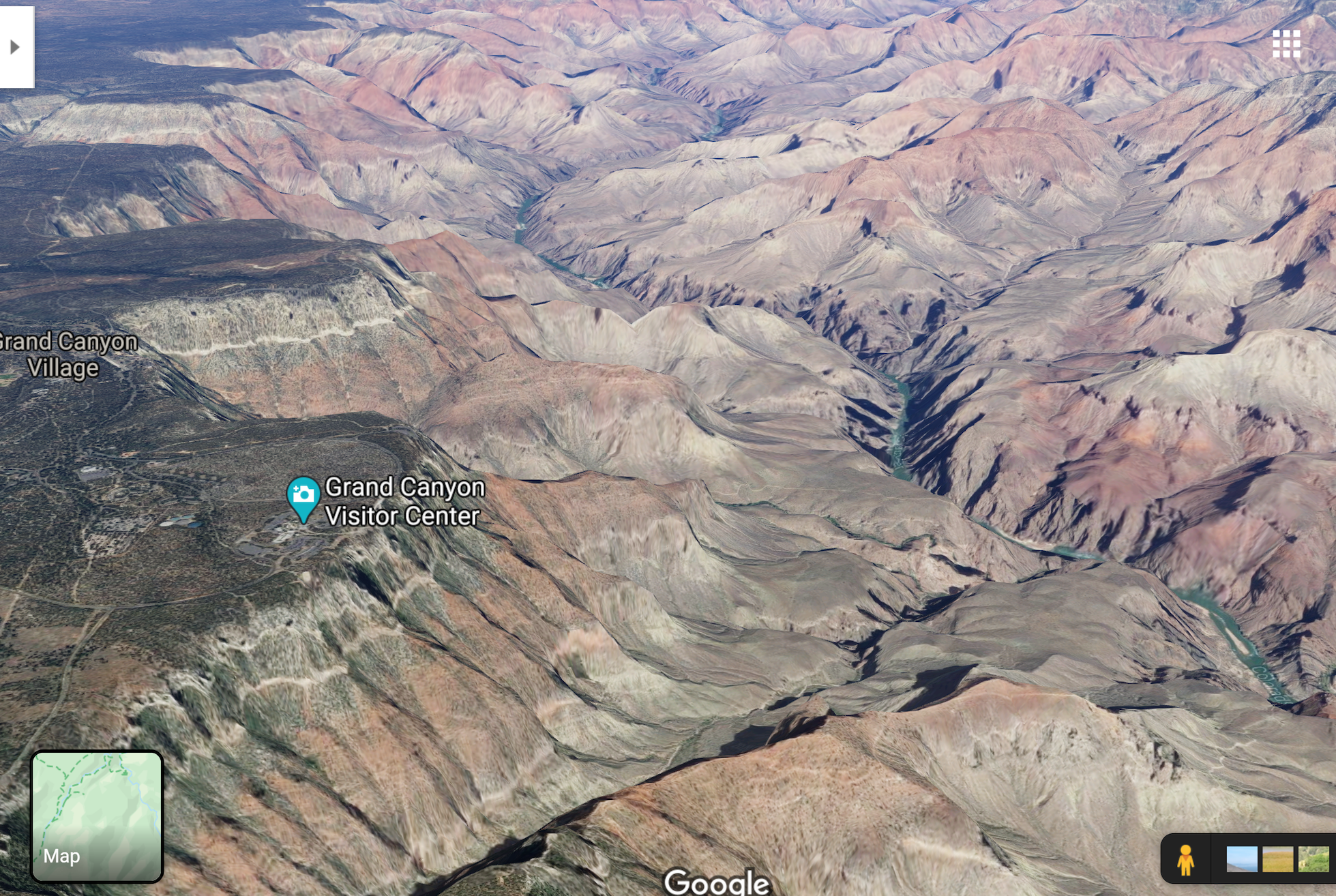}
        \caption{An image from Google Maps of a section of the Grand Canyon corresponding to (36.04, -112.05) latitude and (36.25, -112.3) longitude.}
    \end{subfigure}%
    ~
    \begin{subfigure}[t]{0.23\textwidth}
        \centering
\includegraphics[trim= 20 0 50 20, clip, width=0.95\textwidth]{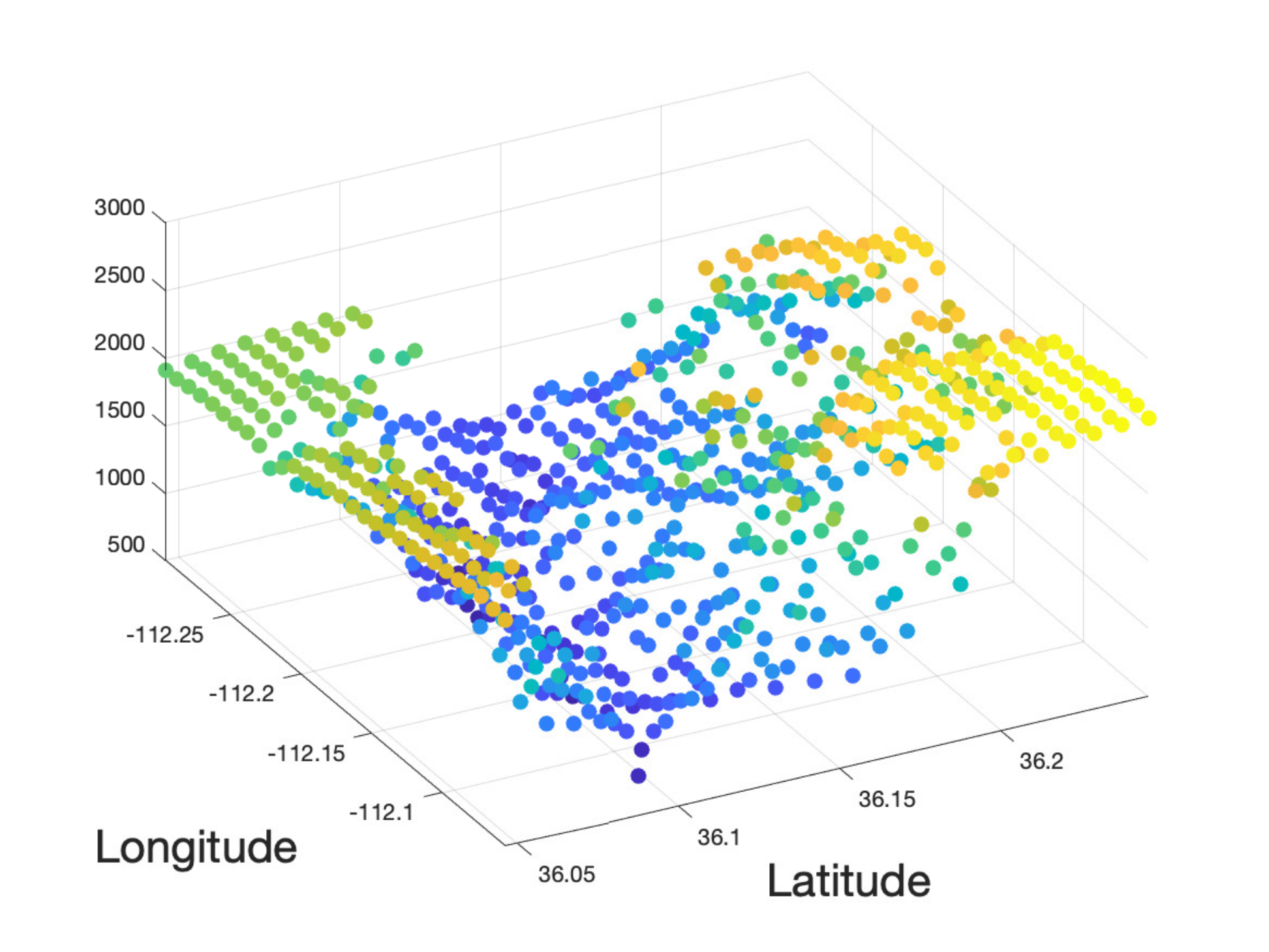}
        \caption{Elevation data ($m=750$) from \cite{becker2009global} for a section of the Grand Canyon between (36.04, -112.05) latitude and (36.25, -112.3) longitude.}
    \end{subfigure}
        ~
    \begin{subfigure}[t]{0.23\textwidth}
        \centering
\includegraphics[trim= 20 0 50 20, clip, width=0.95\textwidth]{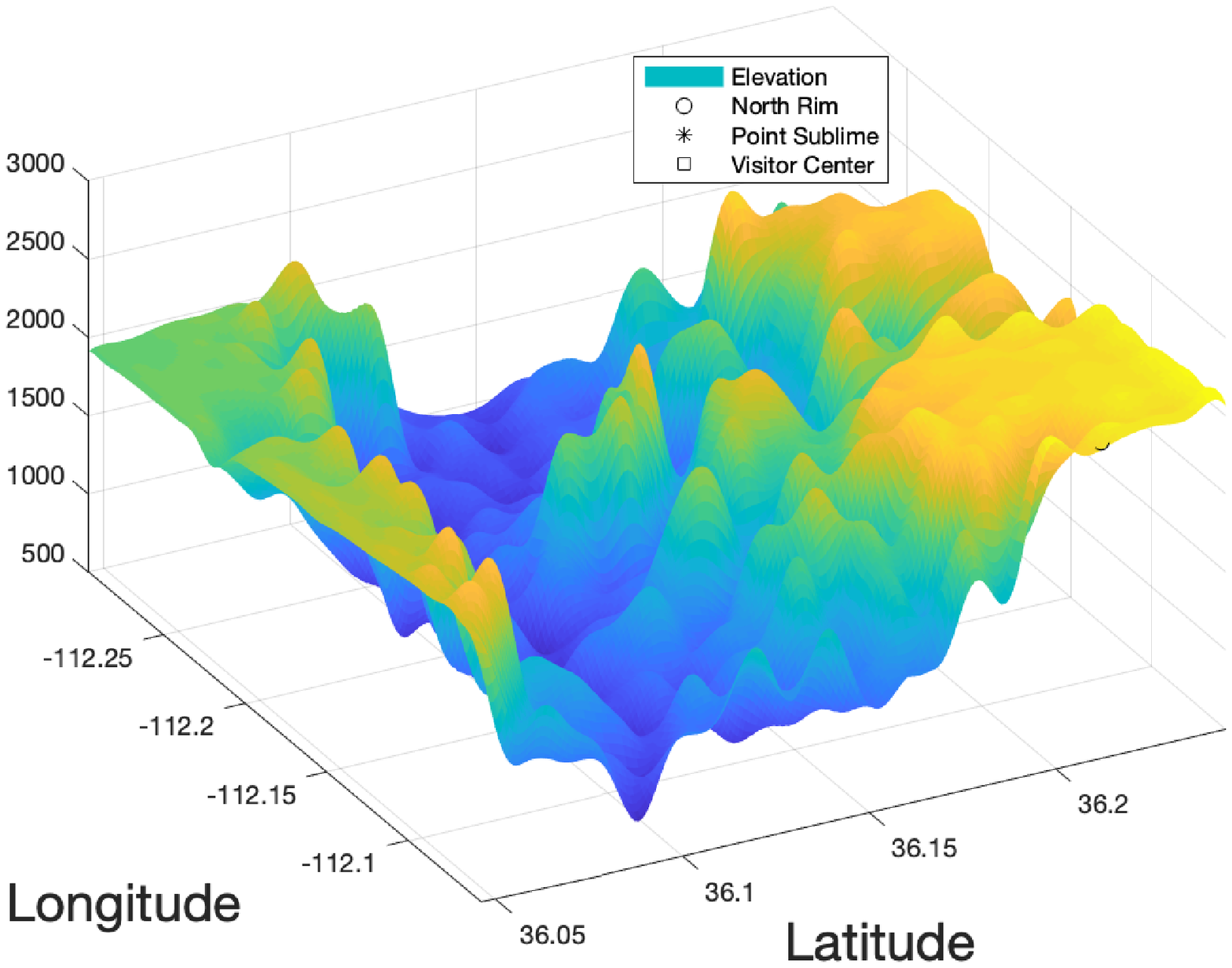}
        \caption{Predictor using a hand-tuned Gaussian kernel trained on the elevation data in (b). The Gaussian predictor poorly represents the sharp edge at the north and south rim.}
    \end{subfigure}
    ~
        \begin{subfigure}[t]{0.23\textwidth}
        \centering
\includegraphics[trim= 20 0 50 20, clip, width=0.95\textwidth]{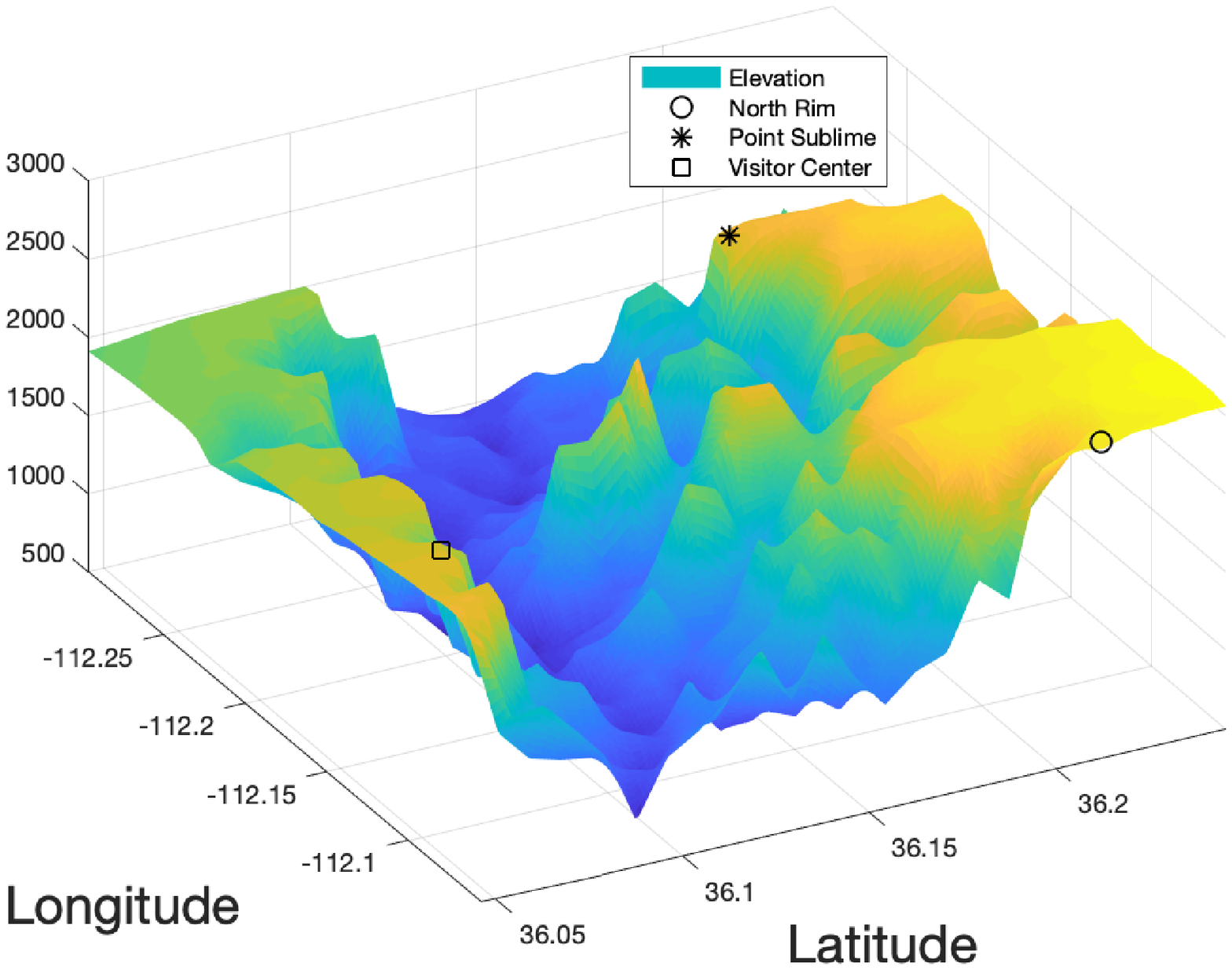}
        \caption{Predictor from Algorithm~\ref{TKL} trained on the elevation data in (b). The TK predictor accurately represent the north and south rims of the canyon.}
    \end{subfigure}
    \caption{Subfigure (a) shows an 3D representation of the section of the Grand Canyon to be fitted. In (b) we plot elevation data of this section of the Grand Canyon. In (c) we plot the predictor for a hand-tuned Gaussian kernel. In (d) we plot the predictor from Algorithm~\ref{TKL} where $d=2$. }\label{fig:GC} \vspace{-5mm}
\end{figure}
\vspace{-2mm}
\subsection{Solving $OPT\_A(P)$ } 
\begin{figure}[t]
    \centering
    \begin{subfigure}[t]{0.23\textwidth}
        \centering
\includegraphics[trim= 20 0 50 20, clip, width=0.8\textwidth]{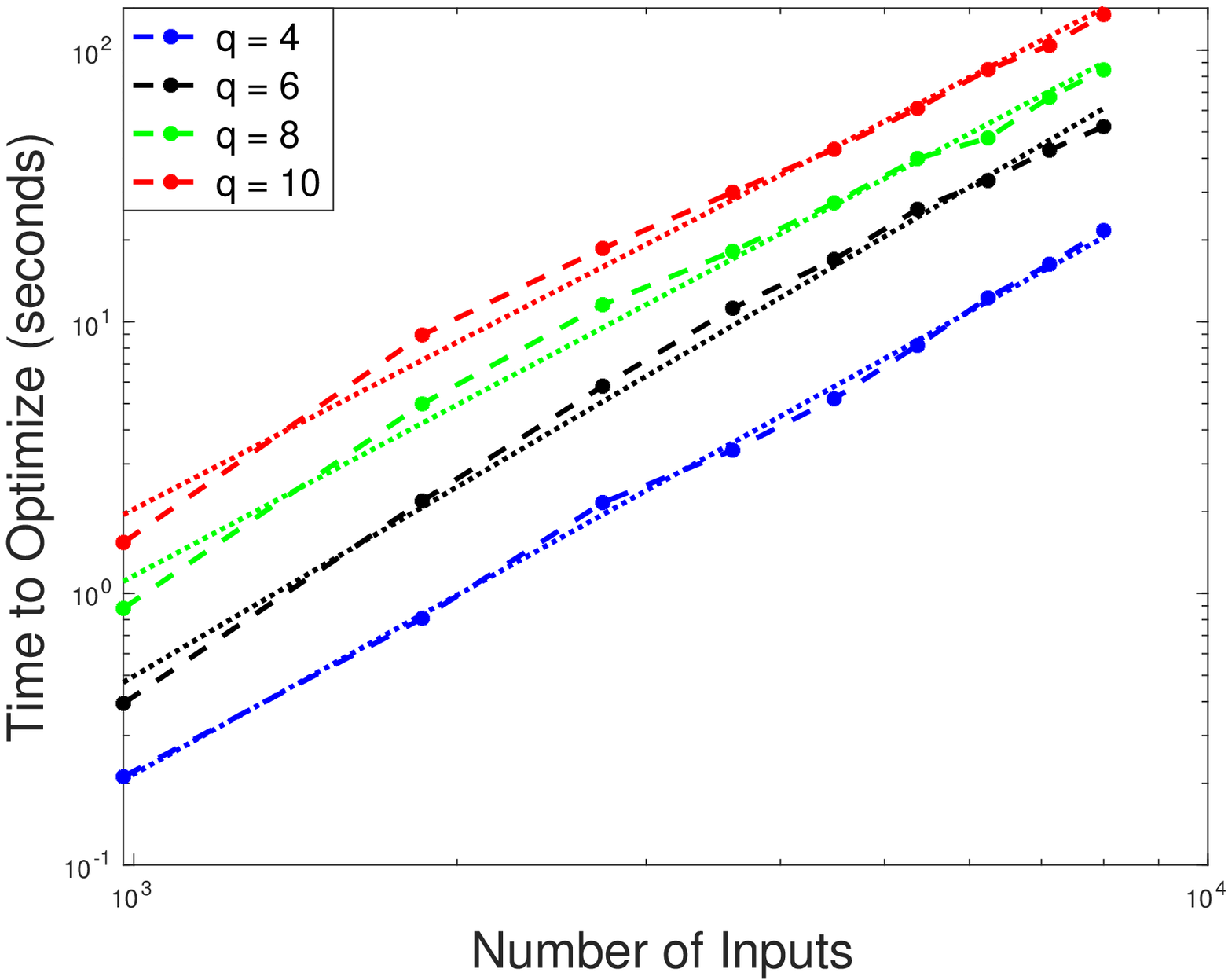}
        \caption{Numerical complexity analysis of TKL for classification versus $m$.}
    \end{subfigure}%
    ~
    \begin{subfigure}[t]{0.23\textwidth}
        \centering
\includegraphics[trim= 20 0 50 20, clip, width=0.8\textwidth]{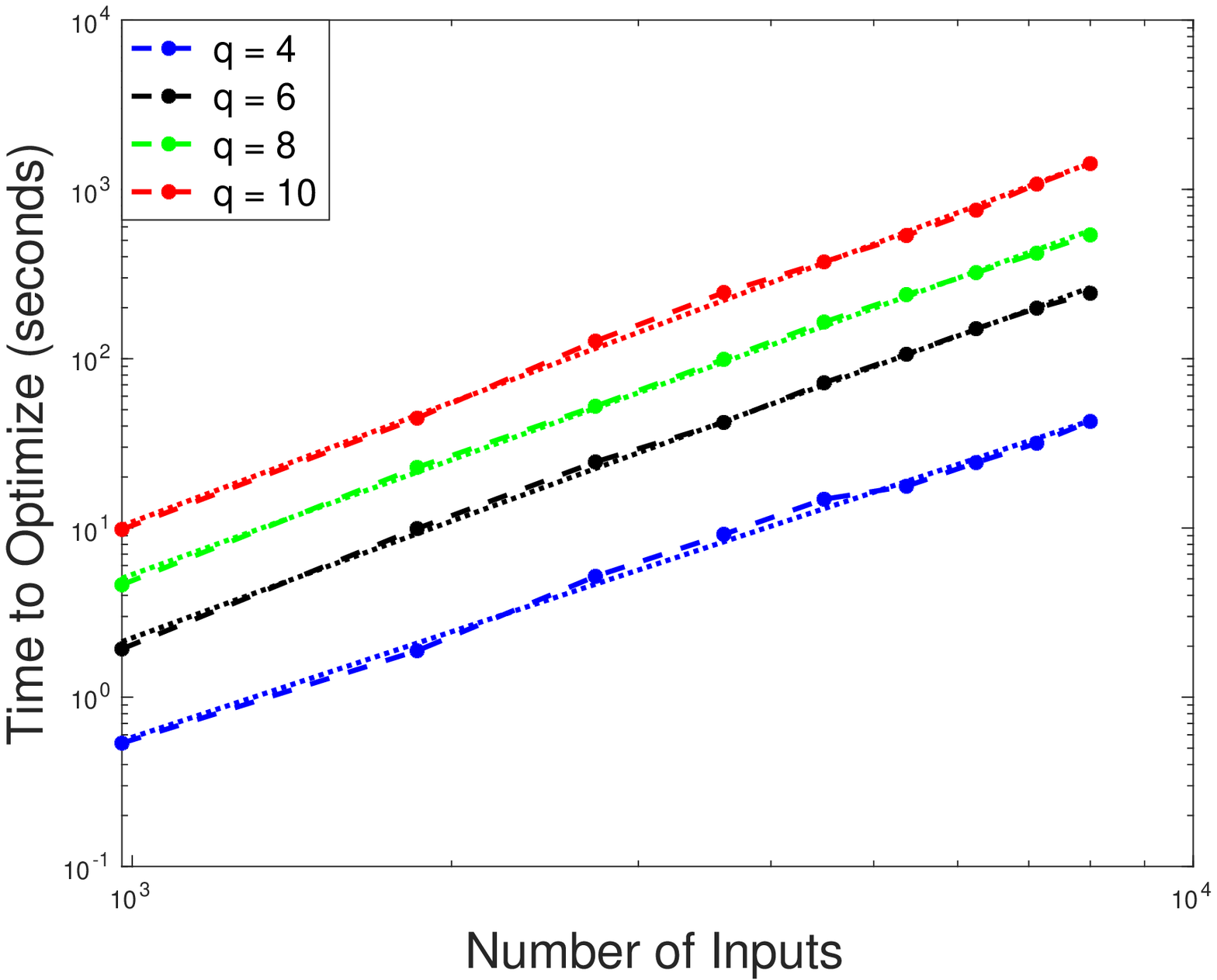}
        \caption{Numerical complexity analysis of TKL for regression versus $m$.}
    \end{subfigure}
        ~
    \begin{subfigure}[t]{0.23\textwidth}
        \centering
\includegraphics[trim= 20 0 50 20, clip, width=0.8\textwidth]{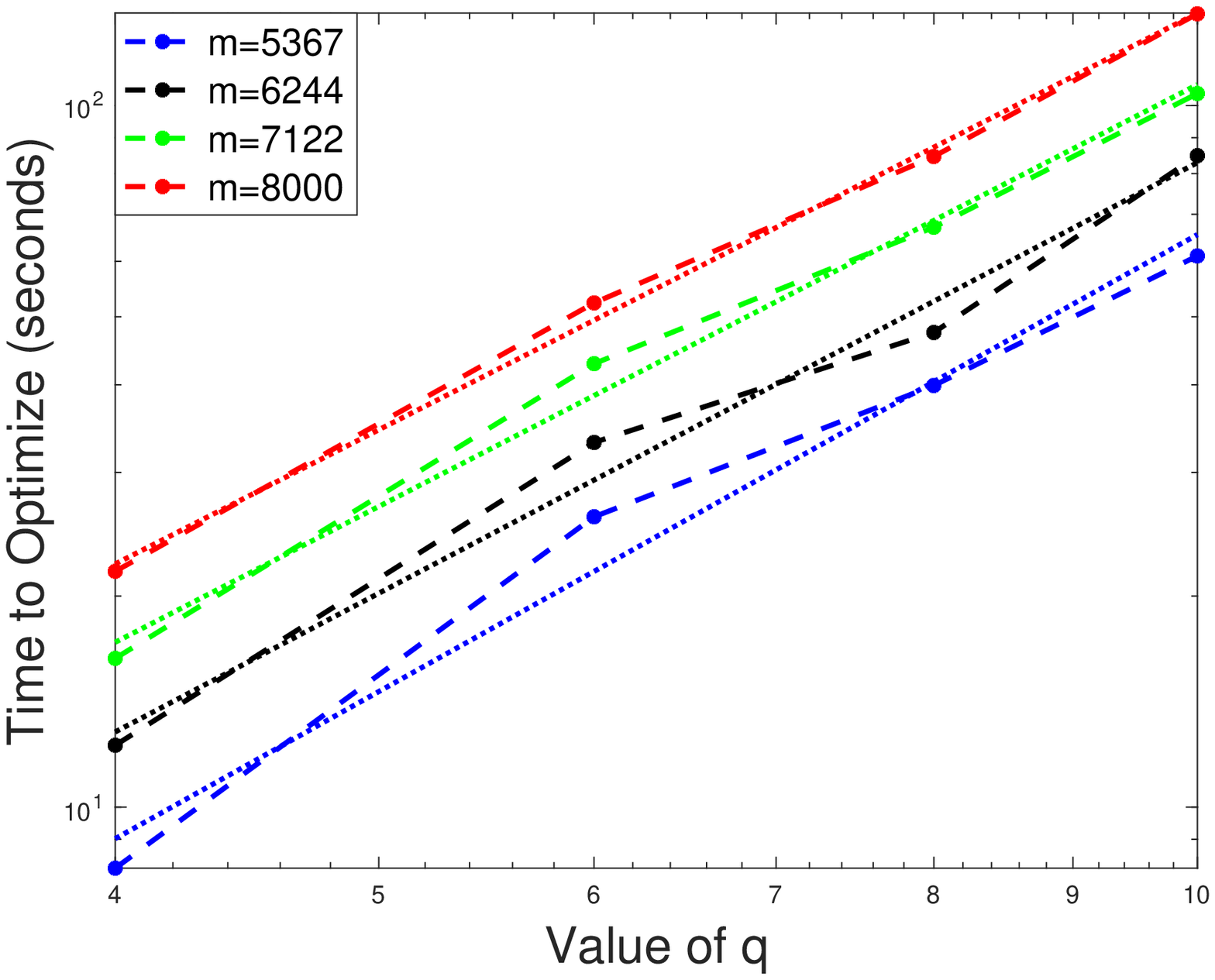}
        \caption{Numerical complexity analysis of TKL for classification versus $q$.}
    \end{subfigure}
    ~
        \begin{subfigure}[t]{0.23\textwidth}
        \centering
\includegraphics[trim= 20 0 50 20, clip, width=0.8\textwidth]{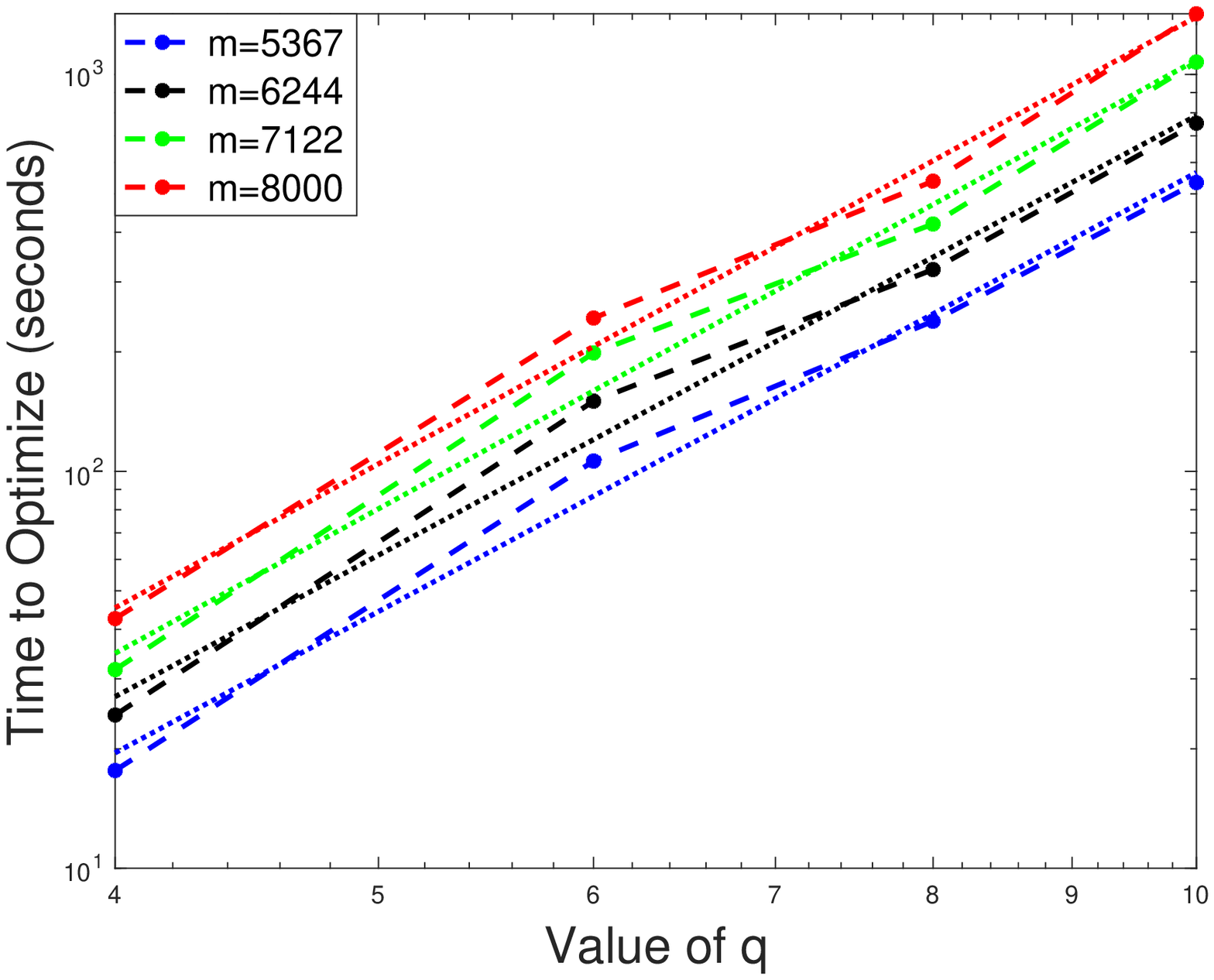}
        \caption{Numerical complexity analysis of TKL for regression versus $q$.}
    \end{subfigure}
    \caption{In (a) and (b) we plot log scale plots of the time taken to optimize TKL as the number of inputs change for $P \in \R^{q\times q}$.  The line of best linear fit is plotted as a dotted line. In (c) and (d) we plot log scale plots of the time taken to optimize TKL as the value of $q$ changes for four different values of $m$. }\label{fig:kernel_complexity} \vspace{-6mm}
\end{figure}

\subsecspace
For a given $P>0$, $OPT\_A(P)$ is a Quadratic Program (QP). General purpose QP solvers as applied to this problem have a worst-case complexity which scales as $O(m^3)$~\cite{ye1989extension} where $m$ is the number of data points. This computational complexity may be improved, however, by noting that the problem formulation is compatible with the representation defined in~\cite{LibSVM} for QPs derived from SVM. In this case, the algorithm in LibSVM~\cite{LibSVM} can reduce the computational burden somewhat. This improved performance is illustrated in Figure~\ref{fig:kernel_complexity} where we observe the achieved complexity scales as $O(m^{2.1})$. Note that for the 2-step algorithm proposed in this manuscript, solving the QP in $OPT\_A(P)$ is significantly slower that solving the Singular Value Decomposition (SVD) required for $OPT\_P(\alpha)$, which is defined in the following subsection. However, the achieved complexity of $O(m^{2.1})$ is also significantly faster than solving the large SDP, as described in~\cite{lanckriet_2004},~\cite{qiu2005multiple}, and~\cite{JMLR}. This complexity comparison will be further discussed in Section~\ref{sec:scalability}.\presecspace
%
%
%
%
%

\subsection{Solving $OPT\_P(\alpha)$} \subsecspace
For a given $\alpha$, $OPT\_P(\alpha)$ is an SDP. Fortunately, however, this SDP is structured so as to admit an analytic solution using the SVD.
To solve $OPT\_P(\alpha)$ we minimize $O(\alpha,P)$ from Eq.~\eqref{OalphaP}
which, as per Corollary~\ref{corr}, is linear in $P$ and can be formulated as
\begin{align}  \label{opt:1}
OPT\_P(\alpha):=\min_{\substack{P \in \R^{q \times q}\\ \text{trace}(P) = q \\ P > 0}}  O(\alpha, P) := \min_{\substack{P \in \R^{q \times q}\\ \text{trace}(P) = q \\ P > 0}}  \text{trace}(C(\alpha)^T P)
\end{align}
where\vspace{-5mm}
\[C_{i,j}(\alpha) = \sum_{k,l = 1}^m (\alpha_k y_k) G_{i,j}(x_k,x_l) (\alpha_l y_l), \qquad G_{i,j}(x,y) := \begin{cases} g_{i,j}(x,y) & \text{if}~ i \leq \frac{q}{2}, j \leq \frac{q}{2} \\ t_{i,j}(x,y) & \text{if}~ i \leq \frac{q}{2}, j > \frac{q}{2} \\ t_{i,j}(y,x) & \text{if} ~i > \frac{q}{2}, j \leq \frac{q}{2} \\ h_{i,j}(x,y) & \text{if}~ i > \frac{q}{2}, j > \frac{q}{2} \end{cases}
\]
and $g$, $t$ and $h$ can be found in Corollary~\ref{corr}.

The following theorem gives an analytic solution for $OPT\_P$ using the SVD.
\begin{thm}
Let $C=V\Sigma V^T$ be the SVD of symmetric $C \in \R^{q \times q}$ and $v$ be the right singular vector corresponding to the minimum singular value of $C$. Then $P^* = q vv^T$ solves $OPT\_P$.
\end{thm} \proofspace
\begin{proof}
Recall $OPT\_P$ has the form $\underset{P \in \R^{q \times q}}{\min}  \text{trace}(C^T P)~~\text{s.t.}~P\geq 0,~\text{trace}(P) = q$. \\
Denote the minimum singular value of $C$ as $\sigma_{\min}(C)$.  Then for any feasible $P\in \mcl X$, by~\cite{fang1994inequalities} we have
\[\text{trace}(C^TP) \geq \sigma_{\min}(C) trace(P) = \sigma_{\min}(C) q. \]
Now consider $P = q vv^T \in \R^{q \times q}$. $P$ is feasible since $P\ge 0$, and $\text{trace}(P) = q$. Furthermore,
\begin{align*}
\text{trace}(C P) ~=~q\,\text{trace}(V\Sigma V^T vv^T)~=~q\,\text{trace}(v^TV \Sigma V^T v) =~q\,\sigma_{\min}(C)
\end{align*}
as desired.\vspace{-2mm}
\end{proof}
Note that the size of the SVD problem in $OPT\_P(\alpha)$ is $q^2$, which increases with the number of features, which is typically relatively small. As a result, we observe that the $OPT\_P$ step of Algorithm~\ref{TKL} is typically less computationally intense than the $OPT\_A$ step.
\presecspace

\section{Complexity and Scalability of the New TK Kernel Learning Algorithm}\label{sec:scalability} \secspace
We consider the computational complexity of Algorithm~\ref{TKL}.
If we define the number of data points used to learn the TK kernel function as $m$ and the size of $P$ as $q \times q$, then we find experimentally that the complexity of Algorithm~\ref{TKL} scales as approximately $O(m^{2.16}q^{2.23})$ for classification and $O(m^{2.24}q^{3.59})$ for regression as can be seen in Fig.~\ref{fig:kernel_complexity}. These results are lower with respect to $m$ than the value of $O(m^{2.6}q^{1.9})$ reported in~\cite{JMLR} for binary classification. The values for classification and regression are both estimated using the data set: Combined Cycle Power Plant (CCPP) in ~\cite{tufekci2014prediction,kaya2012local}, containing 4 features and $m = 9568$ samples. In the case of classification, labels with value greater than or equal to the median of the were relabeled as $1$, and those less than the median were relabeled as $-1$. Note that to study scalability in $q$, we varied the number of features in the dataset - thereby incrementing the size of the matrix $P\in\R^{q\times q}$.

Aside from improved scalability, the overall time required for Algorithm~\ref{TKL} is significantly reduced when compared with the algorithm in~\cite{JMLR}, improving by two orders of magnitude in some cases. This is illustrated for classification using four data sets in Table~\ref{time-table}. This improved complexity is likely due to the lower overhead associated with QP and the SVD.
\presecspace


\begin{table*}[t]
  \caption{We report the mean computation time (in seconds), along with standard deviation, for 30 trials comparing the SDP algorithm in~\cite{JMLR} and the new TKL algorithm on several data sets.  All tests are run on a computer with an Intel i7-5960X CPU at 3.00 GHz with 128 Gb of RAM.}
  \label{time-table}
  \centering
\begin{tabular}{  c |  c c c c } \hline
Method & Liver \cite{UCI}   & Cancer \cite{mangasarian1990pattern}  &  Heart \cite{UCI} & Pima \cite{UCI} \\ \hline
SDP       &  95.75 $\pm$ 2.68 &  636.17 $\pm$ 25.43 & 221.67 $\pm$ 29.63 & 1211.66 $\pm$ 27.01 \\
TKL       &  1.10  $\pm$ 0.24 &  8.20 $\pm$ 0.36 & 3.35 $\pm$ 0.26 & 12.66 $\pm$ 0.44 \\ \hline
\end{tabular}
\end{table*}

\section{Accuracy of the New TK Kernel Learning Algorithm for Regression}\label{sec:8} \secspace
As expected, for classification, the accuracy of the new TK kernel learning algorithm (TKL) is identical to the analysis in~\cite{JMLR}.

For regression, we evaluate the accuracy of TKL when compared to other state of the art machine learning algorithms. Because the set of TK kernels is dense, for classification (as shown in~\cite{JMLR}), TKL outperforms all existing algorithm with respect to TSA. For regression, the appropriate metric is Mean Square Error (MSE). The algorithms used in our comparison are as follows.

\noindent\textbf{[TKL]} Algorithm~\ref{TKL} with $d=1$, $\epsilon = .1$ and we scale the data so that $x_i \in [0,1]^n$, and then select $[a,b] = [0-\delta,1+\delta]^n$, where $\delta>0$ and $C$ are chosen by 5-fold cross-validation;

\noindent\textbf{[SimpleMKL]} We use SimpleMKL \cite{rakotomamonjy_2008} with a standard selection of Gaussian and polynomial kernels with bandwidths arbitrarily chosen between .5 and 10 and polynomial degrees one through three - yielding approximately $13(n+1)$ kernels. We set $\epsilon = .1$ as in TKL and $C$ is chosen by 5-fold cross-validation;

\noindent\textbf{[Neural Net]} We use a 3 layer neural network with 50 hidden layers using MATLABs (\texttt{feedforwardnet}) implementation and stopped learning after the error in a validation set decreased sequentially 50 times.

In Table~\ref{table:Reg}, we see the average MSE on the test set for these three approaches as applied to randomly selected regression benchmark data sets where $n$ is the dimension of the data, $m$ is the number of training data and $m_t$ is the number of testing data points.  In all cases except Forest, [TKL] had both a lower (or comparable) computation time and MSE than both SimpleMKL and Neural Net.
In all cases, the MSE for TKL was significantly lower - illustrating the importance of the density property.

To further illustrate the importance of density property and the TKL framework for practical regression problems, we used elevation data from~\cite{becker2009global} to learn a TK kernel and associated SVM predictor representing the surface of the Grand Canyon in Arizona. This data set is particularly challenging due to the variety of geographical features. The result of the TKL algorithm can be seen in Figure~\ref{fig:GC}(d).
\presecspace

\begin{table*}[t]
  \caption{Mean Square Error comparison for algorithms [TKL], [SimpleMKL] and [Neural Net].  In the data set column $m$ is the number of points in the training data set and $n$ is the number of features.  All tests are run on a computer with an Intel i7-5960X CPU at 3.00 GHz with 128 Gb of RAM.}
  \label{table:Reg}
  \centering \small
\begin{tabular}{  c |  c  c c || c |  c  c c } \hline
Data Set & Method  &   Error   & Time  & Data Set & Method        &   Error                & Time \\ \hline
CCPP \cite{tufekci2014prediction,kaya2012local}        &  TKL    &  9.70   & 1463.8  & Abalone \cite{UCI}             &  TKL    & 3.43    &  522.5  \\
$n$ = 4, $m$ = 8000  &  SimpleMKL      &    13.77      & 26097.1  & $n$ = 8, $m$ = 4000  &  SimpleMKL      &  4.28  &  1185.3 \\
$m_t$ = 1568       &  Neural Net     &    15.00       & 850.4 & $m_t$ = 177       &   Neural Net  &          8.72 & 483.4 \\ \hline
Airfoil  \cite{UCI}     &  TKL    &  1.46    &  92.1  & Forest  \cite{cortez2007data}            &  TKL    &  2.05    &  7.6  \\
$n$ = 5, $m$ = 1300 &  SimpleMKL      &    3.63      &  1025.0 & $n$ = 10, $m$ = 457  &  SimpleMKL      &    2.07      &  0.8 \\
$m_t$ = 203       &   Neural Net  &     4.28      & 61.3   & $m_t$ = 50       &   Neural Net  &     6.40      & 117.7 \\ \hline
\end{tabular}
\vspace{-5mm}
\end{table*}

\section{Conclusion} \secspace
We have extended the TK kernel learning framework to regression problems and proposed a faster algorithm for TK kernel learning which can be used for both classification and regression. The set of TK kernels is tractable, dense, and universal - implying that KL algorithms based on TK kernels are more robust - resulting in higher TSA for classification and lower MSE for regression. These three properties, combined with the improved computational complexity of the new algorithm, has resulted in a kernel learning framework which achieves both lower MSE and computation time when compared to both SimpleMKL and neural networks.


\newpage

\subsubsection*{Broader Impacts}
While machine learning algorithm have become very accurate in recent years, they perform poorly when faced with changes in the underlying process. As evidenced by Covid19, predictive models based on ML algorithms can be brittle~\cite{mims_2020}. The density property of the TK class ensures that the models generated using the algorithms described in this manuscript will be more robust to such changes in environment. Naturally, however, over-reliance on predictive models, without understanding of the process, can lead to negative outcomes, even if the models are robust.

\bibliographystyle{plain}
\bibliography{kernel_methods}
\end{document}